\documentclass[12pt,a4paper]{article}
\pdfoutput=1

\usepackage[utf8]{inputenc}
\usepackage{amsmath}
\usepackage{amsfonts}
\usepackage{amssymb}
\usepackage{algorithmic}
\usepackage{algorithm2e}
\usepackage{cite}
\usepackage{amsthm}
\usepackage{breqn}
\usepackage{titling}
\usepackage{lipsum}
\usepackage{authblk}
\usepackage{graphicx}
\usepackage{afterpage}
\usepackage{comment}
\usepackage{url}

\begin{document}

\title{Victory Probability in the Fire Emblem Arena}
\author{Andrew Brockmann}
\date{August 23, 2018}

\maketitle

\begin{abstract}
We demonstrate how to efficiently compute the probability of victory in Fire Emblem arena battles. The probability can be expressed in terms of a multivariate recurrence relation which lends itself to a straightforward dynamic programming solution. Some implementation issues are addressed, and a full implementation is provided in code.
\end{abstract}

\section{Introduction}

Fire emblem is a series of tactical role-playing video games. Combat is turn-based and alternates between the Player Phase, in which the player is given the chance to move all of his/her units, and the Enemy Phase, in which the enemy AI may do the same. Several common objectives include defeating all enemies, defeating the enemy boss, and seizing the enemy stronghold. Battle continues until the player achieves the map objective or is defeated.

An infrequent but recurring element of Fire Emblem is the arena. Some battle maps feature an arena where the player may optionally send their units to fight. When a unit arrives at the arena, the player must place a wager, after which the player unit engages in single combat against an enemy. If the player wins the battle, he/she obtains money equal to the wager; else, if the player loses or withdraws from the battle, the wager is lost. The arena can be a valuable source of money and experience points. However, repeated failure may result in a net loss of money without experience point gain. Furthermore, defeated player units in Fire Emblem are treated as ``dead'' and cannot be used in future battles---the ``permadeath'' aspect of the series applies to arena battles in many Fire Emblem games.

\afterpage{
\begin{center}
\begin{figure}
\includegraphics[scale=0.8]{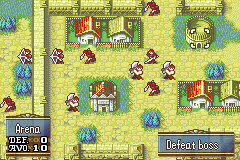}
\includegraphics[scale=0.8]{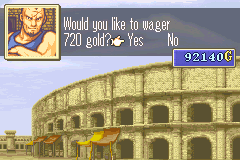}
\caption{The image on the left\cite{fewiki} is from chapter 5 of Fire Emblem: The Sacred Stones. The building in the upper right is an arena. When a unit is sent to the arena, the player is given the option to place a wager and start an arena battle, as seen in the image on the right\cite{rpgclassics}.}
\end{figure}
\end{center}
}

Arena battles are simpler than Fire Emblem overall. Terrain and unit positioning are irrelevant, as are enemy reinforcements and weapon choice---the player and enemy meet face to face and trade blows until one is defeated. Despite this simplicity, it is nontrivial to calculate the player's probability of victory in the arena. Battles can, in principle, continue indefinitely, so the victory probability cannot be computed by simply enumerating all possible battles from start to finish.

This paper demonstrates how to compute the exact victory probability as a straightforward, if somewhat tedious, application of dynamic programming. We begin with a simplified case in which each combatant attacks once per round and neither can perform critical hits. We then extend this result to account for critical hits, and finally deal with the cases where one combatant is fast enough to attack twice per round of combat. In all cases, the victory probability is derived in terms of a multivariate recurrence relation.

\section{Background}

While Fire Emblem gameplay is complicated in general, relatively little background is needed to understand arena battles. Player and enemy units, respectively, sport blue and red color schemes. The battle animation screen displays some key statistics about both combatants (see figure 2 \cite{lp}), with little variation from game to game. At the bottom of the screen are the combatants' hit points, indicating how much damage each unit can take before dying. The sides of the screen offer information about attacks initiated by each unit:
\begin{itemize}
\item \textbf{HIT}: The hit probability of each of the unit's attacks as a percent. (Note: displayed hit rates are inaccurate in most Fire Emblem games. See section 7.2 for more information.)
\item \textbf{DMG}: The damage inflicted by each of the unit's attacks.
\item \textbf{CRT}: The percent probability (conditional on hitting) that each of the unit's attacks will be a \emph{critical hit}. Critical hits inflict thrice the normal damage.
\end{itemize}

\begin{center}
\begin{figure}
\includegraphics[scale=0.8]{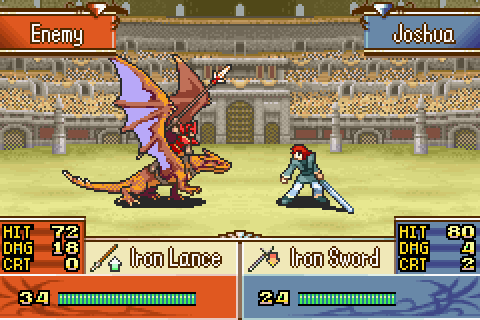}
\caption{An arena battle from Fire Emblem: The Sacred Stones, in which Joshua (player unit) battles an enemy wyvern rider.}
\end{figure}
\end{center}

In Fire Emblem battles, a sufficiently fast unit may attack twice per round of combat while only being attacked once in turn. The speed threshold for performing a \emph{follow-up attack} varies; in recent games, a unit must have an effective speed stat at least 4 or 5 higher than the opponent's in order to perform a follow-up attack. Follow-up attacks are accounted for in battle preview screens but are unmentioned in arena battle screens in most games.

We are now equipped to understand basic arena battles. Combat proceeds according to the following control flow: \\

\begin{algorithmic}[H]
\WHILE{Player and Enemy are both alive}
\STATE Player attacks Enemy
\STATE Enemy attacks Player
\IF{Player is sufficiently fast}
\STATE Player attacks enemy again
\ELSIF{Enemy is sufficiently fast}
\STATE Enemy attacks Player again
\ENDIF
\ENDWHILE
\end{algorithmic}

\section{No Critical Hits or Follow-Up Attacks}

In this simplified special case, each combatant attacks once per round of combat. The only randomness involved is in deciding whether each attack hits. Let $p_1$ and $p_2$, respectively, be the (true) hit probabilities of the player and enemy. Also let $H_1, H_2$ denote the hit points of the player and enemy, and let $d_1, d_2$ be the damage dealt by each player/enemy attack.

Rather than keeping track of the combatants' hit points, it will be more convenient to take note of how many more attacks must hit to fell the player/enemy. For example, in figure 2, Joshua must hit the enemy wyvern rider $\lceil 34/4 \rceil = 9$ times to win. Let $m$ be the number of hits needed to fell the player, and define $n$ similarly for the enemy. Note in particular that
	\[m = \left\lceil \frac{H_1}{d_2} \right\rceil, ~~~~~ n = \left\lceil \frac{H_2}{d_1} \right\rceil.\]
A modicum of caution is needed here: it is possible for $d_1$ and/or $d_2$ to be 0. If neither combatant can hurt the other (i.e. if $d_1 = d_2 = 0$), then the arena battle will never end. Otherwise, if only one unit is able to damage the other, then the battle will be a guaranteed win for that unit. No calculation is needed in these cases.

Now let $A_{m,n}$ be the player's victory probability for a given choice of $m, n < \infty$. We will express $A_{m,n}$ in terms of other $A_{m', n'}$ satisfying:
\begin{itemize}
\item $m' \leq m$
\item $n' \leq n$
\item At least one of $m' < m$ and $n' < n$
\end{itemize}
In other words, we will treat $\left\{A_{m,n}\right\}$ as a multivariate sequence and establish a recurrence relation in which $A_{m,n}$ depends only on ``prior'' terms. The recurrence can be derived with basic probability theory by considering all possible outcomes of the first round of combat. Letting $W$ represent the event that the player wins and using $X$ to denote a possible outcome of the first round, we have:
	\[A_{m,n} := \text{Pr}[W] = \sum_{X} \text{Pr}[X] \cdot \text{Pr}[W | X]. \tag{3.1}\]
For example, one possible outcome $X$ of the first round is that the player and enemy attacks both hit. This happens with probability $\text{Pr}[X] = p_1p_2$. Each combatant will then be one hit closer to death, so the probability of the player winning after this first round is $\text{Pr}[W | X] = A_{m-1,n-1}$.

There are four possible outcomes of the first round, since the player and enemy attack can both independently hit or miss. Using the formula above, we obtain:
	\[A_{m,n} = p_1p_2A_{m-1,n-1} + p_1(1-p_2)A_{m,n-1} + (1-p_1)p_2A_{m-1,n} + (1-p_1)(1-p_2)A_{m,n}.\]
Observe that $A_{m,n}$ appears on both sides of the equality. Expanding $(1-p_1)(1-p_2)$ as $1 - p_1 - p_2 + p_1p_2$ and solving for $A_{m,n}$, we obtain our recurrence relation:
	\[A_{m,n} = \frac{p_1p_2A_{m-1,n-1} + p_1(1-p_2)A_{m,n-1} + (1-p_1)p_2A_{m-1,n}}{p_1 + p_2 - p_1p_2}.\]
To complete the definition of the sequence $\left\{A_{m,n}\right\}$, we must also specify initial values. For example, we should have $A_{m,0} = 1$ whenever $m > 0$---the battle is a guaranteed victory if the player is still alive and the enemy is already dead. We similarly have $A_{0,n} = 0$ when $n > 0$.

Since the recurrence for $A_{m,n}$ includes the term $A_{m-1,n-1}$, it is possible that the recurrence will call on the value $A_{0,0}$, which seems odd because $m = n = 0$ suggests that the player and enemy are both dead. This case can only occur when the player and enemy both deliver the lethal blow in the same round. But the player attacks first during each round, so the player will have dealt the lethal blow first. Therefore, we want $A_{0,0} = 1$. With our initial values accounted for, we can define $\left\{A_{m,n}\right\}$ compactly as follows:
	\[A_{m,n} = \begin{cases}
	{1} & {n \leq 0} \\
	{0} & {m \leq 0, n > 0} \\
	{\frac{p_1p_2A_{m-1,n-1} + p_1(1-p_2)A_{m,n-1} + (1-p_1)p_2A_{m-1,n}}{p_1 + p_2 - p_1p_2}} & {\text{else}} \end{cases}\]
Note that by writing e.g. $n \leq 0$ instead of $n = 0$, we allow for the possibility of negative indices. Thus far, this is impossible---both $m$ and $n$ will be positive at the beginning of each round, and each can then decrease by at most 1 before the next round starts. However, negative indices will become a consideration in the more general cases with critical hits and follow-up attacks, as each unit may take multiple attacks' worth of damage in a single round of combat.

\section{No Follow-Up Attacks}

This section generalizes the previous one by allowing for the possibility of critical hits (while still prohibiting all follow-up attacks). Critical hits change little in principle, although they do make for a more complicated recurrence since there are more possible outcomes of each round of combat. Let $c_1$ and $c_2$, respectively, be the critical hit rates of the player and enemy. Recall that the crit rate is conditional on an attack hitting---therefore, the probability of the player delivering a critical hit is $p_1c_1$, not $c_1$.

Besides the lengthier recurrence, the main difference in this case is that a critical hit deals the damage equivalent of three regular attacks. We must therefore be a little more careful in our definition of $m$ and $n$: for example, we now let $m$ be the number of \emph{regular} attacks needed to fell the player.

The probability formula (3.1) for $A_{m,n}$ from the previous section works here, too. One possible round outcome $X$ is that the player lands a critical hit while the enemy misses. Here, we have:
	\[\text{Pr}[X] = p_1c_1(1-p_2).\]
The probability of the player winning after that particular round outcome would be
	\[\text{Pr}[W | X] = A_{m,n-3},\]
since the enemy takes the equivalent of three regular attacks while failing to deal any damage back to the player.

This time, the recurrence relation will not fit on a single line. For convenience, we will split the terms from the right-hand side of (3.1) into groups, given below:
\begin{itemize}
\item The term corresponding to player and enemy both missing:
	\[(1-p_1)(1-p_2)A_{m,n}\]
\item Terms corresponding to player hitting and enemy missing:
	\[p_1(1-p_2) \left[c_1A_{m,n-3} + (1-c_1)A_{m,n-1}\right]\]
\item Player misses, enemy hits:
	\[(1-p_1)p_2 \left[c_2A_{m-3,n} + (1-c_2)A_{m-1,n}\right]\]
\item Player and enemy both hit:
	\[p_1p_2 \left[c_1c_2A_{m-3,n-3} + c_1(1-c_2)A_{m-1,n-3} + (1-c_1)c_2A_{m-3,n-1} + (1-c_1)(1-c_2)A_{m-1,n-1}\right]\]
\end{itemize}
As in the previous case, we can now sum these terms, set the result equal to $A_{m,n}$, and group the $A_{m,n}$ terms together. We obtain:
\begin{eqnarray*}
(p_1 + p_2 - p_1p_2)A_{m,n} & = & p_1(1-p_2) \left[c_1A_{m,n-3} + (1-c_1)A_{m,n-1}\right] \\
& + & (1-p_1)p_2 \left[c_2A_{m-3,n} + (1-c_2)A_{m-1,n}\right] \\
& + & p_1p_2 \left[c_1c_2A_{m-3,n-3} + c_1(1-c_2)A_{m-1,n-3} \right. \\
& & \left. ~~~~~~ + (1-c_1)c_2A_{m-3,n-1} + (1-c_1)(1-c_2)A_{m-1,n-1}\right]
\end{eqnarray*}
We can finish solving for $A_{m,n}$ by simply dividing through by $(p_1 + p_2 - p_1p_2)$, although this makes it more difficult to write the recurrence in a visually appealing manner.

The initial values for $A_{m,n}$ are precisely the same as in the previous case: we want $A_{m,n} = 1$ whenever $n \leq 0$, and $A_{m,n} = 0$ when $m \leq 0$ and $n > 0$. The value $A_{0,0} = 1$ is again correct because it still only arises in cases where the player lands the lethal blow first.

\section{Enemy Follow-Up Attacks}

The last generalization we'll make in this paper is to account for the cases where one combatant is fast enough to attack twice per round. We first consider follow-up attacks from the enemy because it is actually a little more difficult mathematically to deal with follow-up attacks from the player. When the enemy can perform follow-up attacks, arena battles proceed as follows: in round 1, the player attacks once, and then the enemy attacks twice consecutively. Then begins round 2, in which the player attacks once and the enemy attacks twice consecutively again. As always, battle continues until one unit is defeated.

Let $A_{m,n}^e$ denote the player's victory probability when the enemy can perform follow-up attacks. Similar to the introduction of critical hits, enemy follow-up attacks further complicate the recurrence for $\left\{A_{m,n}^e\right\}$ while changing little about our solution method. Formula (3.1) can again be applied as is. When written out in full, the formula will have the lone term $A_{m,n}^e$ on the left-hand side. Corresponding to the event that all attacks in a given round miss, the right-hand side of (3.1) will contain the term
	\[(1-p_1)(1-p_2)^2A_{m,n}^e.\]
We can again group the $A_{m,n}^e$ terms together by expanding $(1-p_1)(1-p_2)^2A_{m,n}^e$ and moving it to the left-hand side of (3.1). The combined $A_{m,n}^e$ term is
	\[(p_1 + 2p_2 - 2p_1p_2 - p_2^2 + p_1p_2^2)A_{m,n}^e. \tag{5.1}\]
We can solve for $A_{m,n}^e$ by setting (5.1) equal to the sum of all remaining terms on the right-hand side of (3.1). We again group some of these terms together for convenience:
\begin{itemize}
\item The player hits, and the enemy misses both attacks:
	\[p_1(1-p_2)^2 \left[c_1A_{m,n-3}^e + (1-c_1)A_{m,n-1}^e\right] \tag{5.2}\]
\item The player misses, while the enemy hits exactly once:
	\[2(1-p_1)p_2(1-p_2) \left[c_2A_{m-3,n}^e + (1-c_2)A_{m-1,n}^e\right] \tag{5.3}\]
\item The player hits, while the enemy hits exactly once:
	\[2p_1p_2(1-p_2) \left[c_1c_2A_{m-3,n-3}^e + c_1(1-c_2)A_{m-1,n-3}^e\right.\]
	\[~~~~~~~~~~~~~~~~~~~~~~~~~~~~~~~~~~~~~~~~~~~~~~~~~ + \left.(1-c_1)c_2A_{m-3,n-1}^e + (1-c_1)(1-c_2)A_{m-1,n-1}^e\right] \tag{5.4}\]
\item The player misses, while the enemy hits both attacks:
	\[(1-p_1)p_2^2 \left[c_2^2A_{m-6,n}^e + 2c_2(1-c_2)A_{m-4,n}^e + (1-c_2)^2A_{m-2,n}^e\right] \tag{5.5}\]
\item The player hits, and the enemy hits both attacks:
	\[p_1p_2^2 \left[c_1c_2^2A_{m-6,n-3}^e + (1-c_1)c_2^2A_{m-6,n-1}^e\right.\]
	\[~~~~~~~~~~~~~~~~~~~~~~~~~~~~~~~~~~~~~~~~ + \left. 2c_1c_2(1-c_2)A_{m-4,n-3}^e + 2(1-c_1)c_2(1-c_2)A_{m-4,n-1}^e\right.\]
	\[~~~~~~~~~~~~~~~~~~~~~~~~~~~~~~~~~~~~~~~~ + \left. c_1(1-c_2)^2A_{m-2,n-3}^e + (1-c_1)(1-c_2)^2A_{m-2,n-1}^e\right] \tag{5.6}\]
\end{itemize}
The complete recurrence for $A_{m,n}^e$ can be obtained by setting expression (5.1) equal to the sum of expressions (5.2) through (5.6) and then dividing by the coefficient on $A_{m,n}^e$.

The initial values for $A_{m,n}^e$ are again exactly the same as in the previous two cases.

\section{Player Follow-Up Attacks}

Let $A_{m,n}^p$ denote the player's victory probability when the player can perform follow-up attacks. While seemingly similar to the previous case, correctly dealing with player follow-up attacks will require some new tech. This is because the initial value $A_{0,0}^p = 1$ is no longer universally correct.

In the previous cases, all player attacks happened before all enemy attacks in any given round. Thus, if the player and enemy both delivered the lethal blow in the same round, then the player won because he/she dealt the decisive blow first.

But the course of battle is different when the player can perform follow-up attacks: in each round, the player attacks once, then the enemy attacks once, and finally the player attacks again. The enemy's one attack occurs between the player's two attacks. If the player and enemy both deal the lethal blow in the same round, then either combatant could be the winner of the battle.

One way to address this problem is to introduce a multiplier function:
	\[B(m,n) = \begin{cases}
	{0} & {m \leq 0, n > 0} \\
	{1} & {\text{else}} \end{cases}\]
For given $m$ and $n$, this function is just 1 unless the enemy has already won the battle, in which case it takes on a value of 0 instead. The terms of formula (3.1) can be multiplied by appropriate $B(x,y)$ to prevent the player from attacking after being defeated. Thus, we can continue to use the initial value $A_{0,0}^p = 1$.

By way of example, consider the event $X$ in which the player and enemy hit all of their attacks but none are critical hits. Using the previous cases as examples, the term of (3.1) corresponding to this event would be
	\[\text{Pr}[X] \cdot \text{Pr}[W | X] = (p_1^2p_2(1 - c_1)^2(1 - c_2)) \cdot A_{m-1,n-2}^p.\]
But when $m = 1$ and $n = 2$, this yields $\text{Pr}[W | X] = A_{0,0}^p = 1$, which is incorrect because the enemy will have dealt the lethal blow before the player. On the other hand, multiplying by $B(m,n)$---using the values of $m$ and $n$ directly after the enemy's attack---gives us
	\[\text{Pr}[W | X] = B(m-1,n-1) \cdot A_{m-1,n-2}^p,\]
which correctly evaluates to 0 when $m = 1$ and $n = 2$. In cases where the player dies before the follow-up attack, the $B(x,y)$ function ``zeroes out'' his/her win probability. In all other cases, $B(x,y)$ takes on a value of 1 and has no multiplicative effect.

A full recurrence for $A_{m,n}^p$ can be derived in the same way as the previous cases provided that we multiply terms by the appropriate $B(x,y)$ values. There is, however, a more elegant way of solving the player follow-up case by way of reduction to the enemy follow-up case. The trick is to shift the somewhat artificial separation between rounds of combat. The reduction can be made clear by comparing side-by-side the attack flows of the two follow-up cases:
\begin{center}
\begin{tabular}{c|cc}
Round \# & Enemy follow-up & Player follow-up \\ \hline
& Player & Player \\
1 & Enemy & Enemy \\
& Enemy & Player \\ \hline
& Player & Player \\
2 & Enemy & Enemy \\
& Enemy & Player \\ \hline
& Player & Player \\
3 & Enemy & Enemy \\
& Enemy & Player \\ \hline
& Player & Player \\
4 & Enemy & Enemy \\
& $\vdots$ & $\vdots$
\end{tabular}
\end{center}
In the player follow-up case, remove the first player attack and ignore the separations between rounds. The resulting attack flow is then Enemy, Player, Player, Enemy, Player, Player, etc. This is exactly the attack flow from the enemy follow-up case, but with ``Player'' and ``Enemy'' swapped. We can therefore use the enemy follow-up case to compute the \emph{enemy's} probability of victory after the player's first attack. Knowing the enemy's victory probability immediately tells us the player's victory probability too.

Indeed, we can even use formula (3.1) again. This time, though, we need to sum over the possible outcomes $X$ of the player's first attack, instead of the possible outcomes of the first round of combat. There are three possible outcomes for the player's first attack: a miss (probability $(1 - p_1)$), a regular hit (probability $p_1(1 - c_1)$), and a critical hit (probability $p_1c_1$).

After---for example---a player miss on the first attack, the enemy's probability of winning is $A_{n,m}^e$, hence the player's probability of winning is $(1 - A_{n,m}^e)$. (Note that $m$ and $n$ have been swapped; so, too, must the values $p_1, p_2$ and $c_1, c_2$.) Accounting for all three possible outcomes of the initial player attack, formula (3.1) leads us to:
	\[A_{m,n}^p = (1 - p_1)\left(1 - A_{n,m}^e\right) + p_1(1 - c_1)\left(1 - A_{n-1,m}^e\right) + p_1c_1\left(1 - A_{n-3,m}^e\right).\]
This serves as our recurrence relation for the player follow-up case in terms of the enemy follow-up case. Notably, the recurrence above can be used to compute $A_{m,n}^p$ for any choice of $m$ and $n$, so initial values are unnecessary.

\section{Algorithm and Implementation}

Given initial values and a recurrence relation, terms of a sequence can be computed efficiently using dynamic programming. The algorithm is straightforward---our recurrence relations express sequence terms as expressions in lower-indexed sequence terms, so we must fill in the DP table in increasing index order. Pseudocode for computing $A_{m,n}$ is given below:

\begin{center}
\begin{algorithmic}[H]
\STATE Construct an $(m+1) \times (n+1)$ DP table, $A$
\STATE Use initial values of $A_{m,n}$ to fill in some entries
\FOR{$i = 1$ \TO $i = m$}
\FOR{$j = 1$ \TO $j = n$}
\STATE Fill in entry $A[i][j]$ using the recurrence relation
\ENDFOR
\ENDFOR
\RETURN $A[m][n]$
\end{algorithmic}
\end{center}

Values of $A_{m,n}^e$ can be computed in exactly the same way. So, too, can values of $A_{m,n}^p$, using the full self-contained recurrence involving the function $B(x,y)$ (omitted from this paper for brevity). Alternatively, $A_{m,n}^p$ can be computed using the recurrence in terms of $A_{m,n}^e$.

The DP table in our algorithm has $(m+1)(n+1) = O(mn)$ entries. Each of our recurrence relations has constant length, so every entry---initial value or not---takes $O(1)$ time to fill in. Thus, the algorithmic runtime is $O(mn) = O(H_1H_2 / d_1d_2)$.

A GUI implementing this algorithm is provided at the following link:
\begin{verbatim}
https://github.com/aibrockmann/Fire-Emblem-Arena-Probabilities
\end{verbatim}
Several files are available. The first is the source code, a Python script, which can be executed as is if Python 2.7+ is installed. The other files are Windows and Linux executables, created from the source code using PyInstaller.

There are some considerations that arise when implementing this algorithm. Most are ordinary programming considerations. One, however, results from the inaccuracy of displayed hit rates in most Fire Emblem games.

\subsection{Programming Issues}

One programming issue has already been mentioned: if at least one combatant is unable to damage the other, then we will encounter division by 0 when computing $m$ and $n$. This is easily avoided---calculating $m$ and $n$ is unnecessary in these cases. If neither player nor enemy can deal damage, then the battle will be endless. If only one can deal damage, then he/she will win with probability 1.

Another possible point of trouble is that our recurrences may call on terms with negative indices, whereas the indices on our DP table are non-negative. There are several simple ways to get around this issue. One is to change the recurrences themselves so that each index $i$ is replaced with $\max\{i, 0\}$. Another way is to bump negative indices up to 0 before calling on DP table entries. (This is effectively what the provided implementation does.)

Lastly, trouble can arise from inaccuracy in floating point arithmetic. Terms in our recurrences are often products over many non-integer values, so floating point inaccuracies may add up to be appreciable. The provided implementation avoids this problem by storing intermediate values as fractions; only at the end is the exact answer used to produce a decimal approximation, and the exact answer is also provided in fraction form.

\subsection{Fire Emblem True Hit}

As previously mentioned, the displayed hit rates in most Fire Emblem games are incorrect. Our algorithm is still correct---however, one must make sure to convert from displayed hit rates to true hit rates so that the algorithm uses the correct values of $p_1$ and $p_2$.

Displayed hit rates are accurate in the first five Fire Emblem games. For each accuracy check, these games choose a value from $[0, 1, \dots, 99]$ uniformly at random; if the chosen value is less than the displayed hit rate, then the attack hits, else it misses. For a given hit rate $x$, there are precisely $x$ possible values less than $x$, so that the probability of a hit is $x/100$.

Most remaining games, starting with \emph{Binding Blade} and ending with \emph{Awakening}, generate two random numbers (i.e. ``2RN'') per accuracy check. If the \emph{average} of these two numbers is less than the displayed hit, then the attack will hit. The overall effect is that displayed hit rates 50 and higher are understated while those below 50 are overstated. For example, given a displayed hit rate of 99, the only random number combination (out of 10000) whose average is not below 99 is $(99, 99)$. Thus, a displayed hit rate of $99\%$ corresponds to a true hit rate of $99.99\%$. Under the 2RN system, only displayed hit rates of 0 and 100 are accurate.

Research suggests \cite{fe14} that the true hit rates in \emph{Fire Emblem Fates} are different yet. Displayed hit rates below $50\%$ seem to be accurate. For hit rates 50 and above, the game still generates two random numbers, which are averaged and compared to the displayed hit; however, the average in this case seems to be unevenly weighted. In particular, given random numbers $a$ and $b$, \emph{Fates} compares the weighted average $(3a + b)/4$ to the displayed hit rate. This means that displayed hit rates 50 and above are still understated, but the difference is not as large as it is with the old 2RN true hit mechanics.

\section{Extensions}

Our algorithm and implementation correctly provide the player's victory probability in basic arena battles. However, some games feature relevant gameplay elements that we have not taken into account.

For example, we have not taken any offensive skills into account. One such offensive skill is Luna: each attack performed by a unit with Luna has a chance to cut the target's effective defense or resistance in half. The activation rate as a percent is equal to the user's skill stat. Allowing for most offensive skills is simple in principle---we just need to account for more possible round outcomes in formula (3.1). (We would probably also need to index $A_{m,n}$ with the combatants' hit points rather than the number of ``regular'' hits needed to defeat them.) However, the number of terms in our recurrences will grow roughly exponentially in the number of sources of randomness, so the math will become messy very quickly. Correctly accounting for some skills also requires more information than is displayed on screen. For example, the effect of Luna depends on the target's defense or resistance, and this information is not displayed on the arena battle screen in recent Fire Emblem games.

The Fire Emblem series also features special weapons with effects that are not obvious at a glance. ``Brave'' weapons in most games strike twice with each attack, and each strike can independently hit, miss, land a critical hit, or trigger an offensive skill. The Devil Axe bears a chance of dealing damage to the attacker instead of the target. Most special weapons are theoretically simple to account for but significantly complicate the recurrences. Fortunately, this is a minor point---special weapons rarely, if ever, appear in arena battles.

Another relevant gameplay element is, thus far, limited to \emph{Fates}: arena battles may include backup units that attack once per round of combat. As with most offensive skills and special weapons, these dual strikes are straightforward to account for by broadening the set of possible round outcomes.

However, several offensive skills (Sol and Aether) and one special weapon (Nosferatu) can heal the user, and this does change the math in a significant way. Our recurrences express sequence values in terms of lower-indexed sequence values; our dynamic program takes advantage of this monotonicity to compute values in an appropriate order. The possibility of mid-battle healing breaks this monotonicity. It may still be possible to compute $A_{m,n}$ efficiently using modified recurrences, albeit with a very different implementation: the appropriate recurrence could be used to generate $(H_1 + 1)(H_2 + 1)$ linearly independent equations in the $A_{i,j}$, and the resulting system could be solved using Gaussian elimination. Under this modified implementation, $H_1$ and $H_2$ should be the combatants' \emph{maximum} (rather than current) hit points, while the indices $m$ and $n$ should denote the combatants' current hit points. Regardless of any healing, a unit's current hit points cannot rise above maximum; thus, the system generated in this way would have $O(H_1H_2)$ equations.

It might be possible to find a closed form for $A_{m,n}$ in general. However, work toward a closed form using generating functions has not proved fruitful, even in simplified special cases of the arena problem.

\end{document}